# Sensitivity Maps of the Hilbert-Schmidt Independence Criterion

Adrián Pérez-Suay and Gustau Camps-Valls

Image Processing Laboratory (IPL), Universitat de València, Spain

Catedrático A. Escardino, 46980 Paterna, València (Spain)

**Abstract**

Kernel dependence measures yield accurate estimates of nonlinear relations between random variables, and they are also endorsed with solid theoretical properties and convergence rates. Besides, the empirical estimates are easy to compute in closed form just involving linear algebra operations. However, they are hampered by two important problems: the high computational cost involved, as two kernel matrices of the sample size have to be computed and stored, and the interpretability of the measure, which remains hidden behind the implicit feature map. We here address these two issues. We introduce the Sensitivity Maps (SMs) for the Hilbert-Schmidt independence criterion (HSIC). Sensitivity maps allow us to explicitly analyze and visualize the relative relevance of both examples and features on the dependence measure. We also present the randomized HSIC (RHSIC) and its corresponding sensitivity maps to cope with large scale problems. We build upon the framework of random features and the Bochner's theorem to approximate the involved kernels in the canonical HSIC. The power of the RHSIC measure scales favourably with the number of samples, and it approximates HSIC and the sensitivity maps efficiently. Convergence bounds of both the measure and the sensitivity map are also provided. Our proposal is illustrated in synthetic examples, and challenging real problems of dependence estimation, feature selection, and causal inference from empirical data.

## 1 Introduction

The problem of estimating statistical dependencies between random variables is ubiquitous in Science and Engineering, and the basis to discover causal relations from empirical data. Many methods exist to this purpose. Very often one traditionally resorts to Pearson's correlation, but the measure can only identify linear associations between random variables. Other measures of dependence, such as the Spearman's rank or the Kendall's tau criteria, assume monotonically increasing variable relations, and can be better suited in problems exhibiting such relations. All of them, however, can be computed for pairs of variables only, and thus the multidimensional problem of dependence estimation is tackled by repeating the test for all pairwise combinations, and then summarizing the 'dependence matrix' into an *ad hoc* overall statistic.

In recent years, we have witnessed the introduction of an increasing amount of nonlinear dependence measures. Among the vast amount of criteria, kernel dependence methods exhibit some good properties (Gretton et al., 2005a). They typically reveal 1) good robustness properties in high dimensional and low number of samples settings; 2) criteria are not restricted to estimate pairwise dependencies, but capture higher-order relations between (multidimensional) random variables; 3) the empirical estimates are very simple to implement in closed form and only involve kernel matrix computation and linear algebra operations; 4) there is a well-founded theoretical background to study and characterize them, and fast converge rates to the true measure can be derived; and 5) one can actually derive $p$-values associated to the empirical measure. In this paper, we focus on improving the family of kernel dependence estimates in terms of *computational efficiency* and *interpretability*.

The principle underlying kernel-based dependence estimation is to define covariance and cross-covariance operators in reproducing kernel Hilbert spaces (RKHS) (Schölkopf and Smola, 2002), and derive statistics from these operators capable of measuring dependence between functions therein. In (Bach and Jordan, 2002) the largest singular value of the kernel canonical correlation analysis (KCCA) –which uses both covariance and cross-covariances– was used as a statistic to test independence. Later, in (Gretton et al., 2005c), the constrained covariance (COCO) statistic was proposed, which uses the largest singular value of the cross-covariance operator: high efficiency was obtained with virtually no regularization needed. A variety of empirical kernel quantities derived from bounds on the mutual information that hold near independence were also proposed: namely the kernel Generalised Variance (kGV) and the Kernel Mutual Information (kMI) (Gretton et al., 2003, 2005a). Among the most interesting kernel dependence methods, we find the Hilbert-Schmidt Independence Criterion (HSIC) (Gretton et al., 2005b). The method consists of measuring cross-covariances in a proper RKHS, and generalizes several measures, such as COCO, by using the entire spectrum of the cross-covariance operator, not just the largest singular value. The HSIC empirical estimator is very easy to compute, has good theoretical properties (Gretton et al., 2005b,a), and yields very good results in practice, e.g. for ranking (Song et al., 2007b), clustering (Song et al., 2007c), and feature selection from satellite images (Camps-Valls et al., 2010) and gene expression (Song et al., 2007a).

Kernel dependence estimates such as HSIC however face two main challenges: (1) the measure is hardly interpretable in geometric terms, as it is based on implicit mappings reproduced via reproducing kernel functions; and (2) the method scales poorly with the number of examples, as it involves computing and storing kernel matrices of the sample size. We will tackle these two important limitations in this paper, illustrating the methodology for the particular case of HSIC. Specifically, the contributions are summarized as follows:

- In order to analyze and visualize the kernel dependence measure, we propose to derive *sensitivity maps* (SMs) of the estimate. Sensitivity maps allow us to explicitly analyze and visualize the relative relevance of *both* examples and features on the dependence measure (Kjems et al., 2002). Our inspiration is a probabilistic approach to derive sensitivity maps for Support Vector Machines (SVM) in neuroimage applications (Rasmussen et al., 2011), which has been recently extended to the field of Gaussian Processes (GPs) visualization in geoscience problems (Camps-Valls et al., 2015; Blix et al., 2015). In both cases, the goal was to study the sensitivity (relevance, impact) of features on the learned *supervised* model. In our case, however, we deal with the more challenging *unsupervised* scenario of scrutinizing kernel-based dependence measures. For this, we develop the SMs to visualize and study HSIC dependence measure quantitatively. We will show that the SMs provide a *vector field* that allows us to identify both examples and features most affecting the measure of dependence.

- In order to alleviate the high computational burden involved in both HSIC and its SM, we here introduce the randomized HSIC (RHSIC), and derive an efficient SM that still preserves the appealing closed-form computation property. Essentially, we replace the involved kernels in HSIC by explicit mappings generated via linear projections on random features. This approximation builds upon the framework of random features originally introduced in (Rahimi and Recht, 2007, 2009) and the Bochner's theorem (Reed and Simon, 1981; Rudin, 1987). We want to highlight that introducing the RHSIC is not incidental, but capitalizes on the fact that still permits to derive sensitivity maps in a very efficient, closed-form manner.

The remainder of the paper is organized as follows. Section 2 fixes notation, briefly introduces the HSIC estimate, and presents the randomized HSIC for computational efficiency. We also discuss on the computational gain and on the convergence rates for the estimate, the decision threshold and the associated $p$-values. Section 3 introduces the sensitivity maps for both HSIC and its randomized version. Section 4 shows experiments of the performance of RHSIC and the properties of the sensitivity maps. In particular, we give empirical evidence of performance of the sensitivity maps in both synthetic examples and challenging real problems of dependence estimation, feature ranking, and causal inference from empirical data. Section 5 concludes this paper with some remarks and future research lines. Some theoretical properties of convergence of the randomized measure and its sensitivity are given in the Appendix A.

## 2 Efficient HSIC dependence estimation

To fix notation, let us consider two spaces $\mathcal{X} \subseteq \mathbb{R}^{d_x}$ and $\mathcal{Y} \subseteq \mathbb{R}^{d_y}$, on which we jointly sample observation pairs $(\mathbf{x}, \mathbf{y})$ from distribution $\mathbb{P}_{\mathbf{xy}}$. The covariance matrix can be defined as

$$\mathcal{C}_{\mathbf{xy}} = \mathbb{E}_{\mathbf{xy}}(\mathbf{x}\mathbf{y}^\top) - \mathbb{E}_{\mathbf{x}}(\mathbf{x})\mathbb{E}_{\mathbf{y}}(\mathbf{y}^\top), \tag{1}$$

where $\mathbb{E}_{\mathbf{xy}}$ is the expectation with respect to $\mathbb{P}_{\mathbf{xy}}$, $\mathbb{E}_{\mathbf{x}}$ is the expectation with respect to the marginal distribution $\mathbb{P}_{\mathbf{x}}$ (hereafter, we assume that all these quantities exist), and $\mathbf{y}^\top$ is the transpose of $\mathbf{y}$. The covariance matrix encodes all first order dependencies between the random variables. A statistic that efficiently summarizes the content of this matrix is its Hilbert-Schmidt norm.

The square of this norm is equivalent to the squared sum of its eigenvalues $\gamma_i$:

$$\|\mathcal{C}_{\mathbf{xy}}\|^2_{\text{HS}} = \sum_i \gamma_i^2. \tag{2}$$

This quantity is zero if and only if there exists no first order dependence between $\mathbf{x}$ and $\mathbf{y}$. Note that the Hilbert Schmidt norm is limited to the detection of second order relations, and thus more complex (higher-order effects) cannot be captured.

## 2.1 Kernel dependence estimation

Let us define a (possibly non-linear) mapping $\boldsymbol{\phi} : \mathcal{X} \to \mathcal{F}$ such that the inner product between features is given by a positive definite (p.d.) kernel function $K_x(\mathbf{x}, \mathbf{x}') = \langle \boldsymbol{\phi}(\mathbf{x}), \boldsymbol{\phi}(\mathbf{x}') \rangle$. The feature space $\mathcal{F}$ has the structure of a reproducing kernel Hilbert space (RKHS). Let us now denote another feature map $\boldsymbol{\psi} : \mathcal{Y} \to \mathcal{G}$ with associated p.d. kernel function $K_y(\mathbf{y}, \mathbf{y}') = \langle \boldsymbol{\psi}(\mathbf{y}), \boldsymbol{\psi}(\mathbf{y}') \rangle$. Then, the cross-covariance operator between these feature maps is a linear operator $\mathcal{C}_{\mathbf{xy}} : \mathcal{G} \to \mathcal{F}$ such that $\mathcal{C}_{\mathbf{xy}} = \mathbb{E}_{\mathbf{xy}}[(\phi(\mathbf{x}) - \mu_x) \otimes (\psi(\mathbf{y}) - \mu_y)]$, where $\otimes$ is the tensor product, $\mu_x = \mathbb{E}_{\mathbf{x}}[\boldsymbol{\phi}(\mathbf{x})]$, and $\mu_y = \mathbb{E}_{\mathbf{y}}[\boldsymbol{\psi}(\mathbf{y})]$. See more details in Baker (1973); Fukumizu et al. (2004). The squared norm of the cross-covariance operator, $\|\mathcal{C}_{\mathbf{xy}}\|^2_{\text{HS}}$, is called the Hilbert-Schmidt Independence Criterion (HSIC) and can be expressed in terms of kernels Gretton et al. (2005b). Given the sample datasets $\mathbf{X} \in \mathbb{R}^{n \times d_x}$, $\mathbf{Y} \in \mathbb{R}^{n \times d_y}$, with $n$ pairs $\{(\mathbf{x}_1, \mathbf{y}_1), \ldots, (\mathbf{x}_n, \mathbf{y}_n)\}$ drawn from the joint $\mathbb{P}_{\mathbf{xy}}$, an empirical estimator of HSIC is Gretton et al. (2005b):

$$\text{HSIC}(\mathcal{F}, \mathcal{G}, \mathbb{P}_{\mathbf{xy}}) = \frac{1}{n^2}\text{Tr}(\mathbf{K}_x\mathbf{H}\mathbf{K}_y\mathbf{H}) = \frac{1}{n^2}\text{Tr}(\mathbf{H}\mathbf{K}_x\mathbf{H}\ \mathbf{K}_y), \tag{3}$$

where $\text{Tr}(\cdot)$ is the trace operation (the sum of the diagonal entries), $\mathbf{K}_x$, $\mathbf{K}_y$ are the kernel matrices for the input random variables $\mathbf{x}$ and $\mathbf{y}$, respectively, and $\mathbf{H} = \mathbf{I} - \frac{1}{n}\mathbb{1}\mathbb{1}^\top$ centers the data in the feature spaces $\mathcal{F}$ and $\mathcal{G}$, respectively.

## 2.2 The randomized HSIC

An outstanding result in the recent kernel methods literature makes use of a classical definition in harmonic analysis to improve approximation and scalability Rahimi and Recht (2007, 2009). The Bochner's theorem (Reed and Simon, 1981; Rudin, 1987) states that a continuous kernel $K(\mathbf{x}, \mathbf{x}') = K(\mathbf{x} - \mathbf{x}')$ on $\mathbb{R}^d$ is positive definite (p.d.) if and only if $K$ is the Fourier transform of a non-negative measure. If a shift-invariant kernel $K$ is properly scaled, its Fourier transform $p(\mathbf{w})$ is a proper probability distribution. This property is used to approximate kernel functions and matrices with linear projections on a number of $D$ random features, as follows:

$$K(\mathbf{x}, \mathbf{x}') = \int_{\mathbb{R}^d} p(\mathbf{w}) e^{-\mathbb{i}\mathbf{w}^\top(\mathbf{x}-\mathbf{x}')} d\mathbf{w} \approx \sum\nolimits_{i=1}^{D} \tfrac{1}{D} e^{-\mathbb{i}\mathbf{w}_i^\top \mathbf{x}} e^{\mathbb{i}\mathbf{w}_i^\top \mathbf{x}'}$$

where $p(\mathbf{w})$ is set to be the inverse Fourier transform of $K$, $\mathbb{i} = \sqrt{-1}$, and $\mathbf{w}_i \in \mathbb{R}^d$ is randomly sampled from a data-independent distribution $p(\mathbf{w})$ Rahimi and Recht (2009). Note that we can define a $D$-dimensional *randomized* feature map $\mathbf{z}(\mathbf{x}) : \mathbb{R}^d \to \mathbb{C}^D$, which can be explicitly constructed as $\mathbf{z}(\mathbf{x}) :=$

$[\exp(\mathbb{i}\mathbf{w}_1^\top\mathbf{x}), \ldots, \exp(\mathbb{i}\mathbf{w}_D^\top\mathbf{x})]^\top$. Other definitions are possible: one could for instance expand the exponentials in pairs $[\cos(\mathbf{w}_i^\top\mathbf{x}), \sin(\mathbf{w}_i^\top\mathbf{x})]$, but this increases the mapped data dimensionality to $\mathbb{R}^{2D}$, while approximating exponentials by $[\cos(\mathbf{w}_i^\top\mathbf{x}+b_i)]$, where $b_i \sim \mathcal{U}(0, 2\pi)$, is more efficient (still mapping to $\mathbb{R}^D$) but has revealed less accurate, as studied in (Sutherland and Schneider, 2015).

In matrix notation, given $n$ data points, the kernel matrix $\mathbf{K} \in \mathbb{R}^{n\times n}$ can be approximated with the explicitly mapped data, $\mathbf{Z} = [\mathbf{z}_1 \cdots \mathbf{z}_n]^\top \in \mathbb{R}^{n\times D}$, and will be denoted as $\hat{\mathbf{K}} \approx \mathbf{Z}\mathbf{Z}^\top$. This property can be used to approximate any shift-invariant kernel. For instance, the familiar squared exponential (SE) Gaussian kernel $K(\mathbf{x}, \mathbf{x}') = \exp(-\|\mathbf{x}-\mathbf{x}'\|^2/(2\sigma^2))$ can be approximated using $\mathbf{w}_i \sim \mathcal{N}(\mathbf{0}, \sigma^{-2}\mathbf{I})$, $1 \leq i \leq D$. We here introduce the randomized HSIC (RHSIC) for fast dependence estimation in large-scale problems. We first define two explicit mapping functions over random features to approximate the two kernel matrices involved in HSIC. For the sake of simplicity, Gaussian kernel matrices $\mathbf{K}_x$ and $\mathbf{K}_y$ are then approximated using complex exponentials of projected data matrices, $\mathbf{Z}_x = \exp(\mathbb{i}\mathbf{X}\mathbf{W}_x)/\sqrt{D_x} \in \mathbb{C}^{n\times D_x}$ and $\mathbf{Z}_y = \exp(\mathbb{i}\mathbf{Y}\mathbf{W}_y)/\sqrt{D_y} \in \mathbb{C}^{n\times D_y}$, where $\mathbf{W}_x \in \mathbb{R}^{d_x\times D_x}$ and $\mathbf{W}_y \in \mathbb{R}^{d_y\times D_y}$, both drawn from a Gaussian $\mathbf{W}_x = [\mathbf{w}_1^x|\cdots|\mathbf{w}_{D_x}^x]$, and $\mathbf{W}_y = [\mathbf{w}_1^y|\cdots|\mathbf{w}_{D_y}^y]$.

Since we are working with explicit mappings, centering the points in the complex feature space can be done efficiently because the $n\times n$ centering matrix $\mathbf{H}$ is no longer involved. We will denote the centered matrices with $\tilde{\mathbf{Z}}_x$ and $\tilde{\mathbf{Z}}_y$. Now, plugging the corresponding kernel approximations, $\hat{\mathbf{K}}_x = \tilde{\mathbf{Z}}_x\tilde{\mathbf{Z}}_x^\top$ and $\hat{\mathbf{K}}_y = \tilde{\mathbf{Z}}_y\tilde{\mathbf{Z}}_y^\top$, into (3), and after manipulating terms, we obtain

$$\text{RHSIC}(\mathcal{F}, \mathcal{G}, \mathbb{P}_{\mathbf{xy}}) = \frac{1}{n^2}\Re(\text{Tr}(\tilde{\mathbf{Z}}_x^\top\tilde{\mathbf{Z}}_y\tilde{\mathbf{Z}}_y^\top\tilde{\mathbf{Z}}_x)), \tag{4}$$

which corresponds to Hilbert-Schmidt norm of the randomized cross-covariance operator, $\hat{\mathcal{C}}_{\mathbf{xy}} = \tilde{\mathbf{Z}}_x^\top\tilde{\mathbf{Z}}_y \in \mathbb{R}^{D_x\times D_y}$, which can be computed explicitly and just once. The computational complexity of RHSIC reduces considerably over the original HSIC. A naive implementation of HSIC runs $\mathcal{O}(n^2)$, while the proposed RHSIC cost is $\mathcal{O}(nD^2)$, where $D = D_x = D_y$, since computing matrices $\mathbf{Z}$ only involves matrix multiplications and exponentials.

We would like to emphasize that the proposed randomization of HSIC has some additional good properties: first, it does not require to store the kernel matrix in memory which can be infeasible for large scale problems, and second, and more importantly, having the approximating matrices explicitly allows us to derive sensitivity maps in closed-form (as will be shown in the next section).

## 2.3 Setting the decision threshold

An important issue in kernel-based measures of independence is how to set the decision threshold. This is actually even of higher importance in RHSIC given that the number of random features impact the statistic estimate distribution, and hence setting an appropriate decision threshold. In Song et al. (2012), several statistical tests of independence based on the empirical HSIC estimator (3) were revised. The test should discern between the null hypothesis $\text{H}_0 : \mathbb{P}_{\mathbf{xy}} = \mathbb{P}_{\mathbf{x}}\mathbb{P}_{\mathbf{y}}$ (factorization means independence), and the alternative hypothesis $\text{H}_1 : \mathbb{P}_{\mathbf{xy}} \neq \mathbb{P}_{\mathbf{x}}\mathbb{P}_{\mathbf{y}}$. This is done by comparing the test statistic HSIC (or RHSIC) with a given threshold[1]. Among the possibilities to define such a threshold over the HSIC estimate, a reasonable one is to approximate the null distribution as a two-parameter gamma distribution, as suggested in Johnson et al. (1994):

$$\widehat{\mathrm{HSIC}} \sim \frac{x^{a-1}e^{-x/b}}{nb^a\Gamma(a)}, \tag{5}$$

where $a = \mathbb{E}[\widehat{\mathrm{HSIC}}]^2/\mathbb{V}[\widehat{\mathrm{HSIC}}]$ and $b = \mathbb{V}[\widehat{\mathrm{HSIC}}]/\mathbb{E}[\widehat{\mathrm{HSIC}}]$, whose detailed expressions can be found in Song et al. (2012).

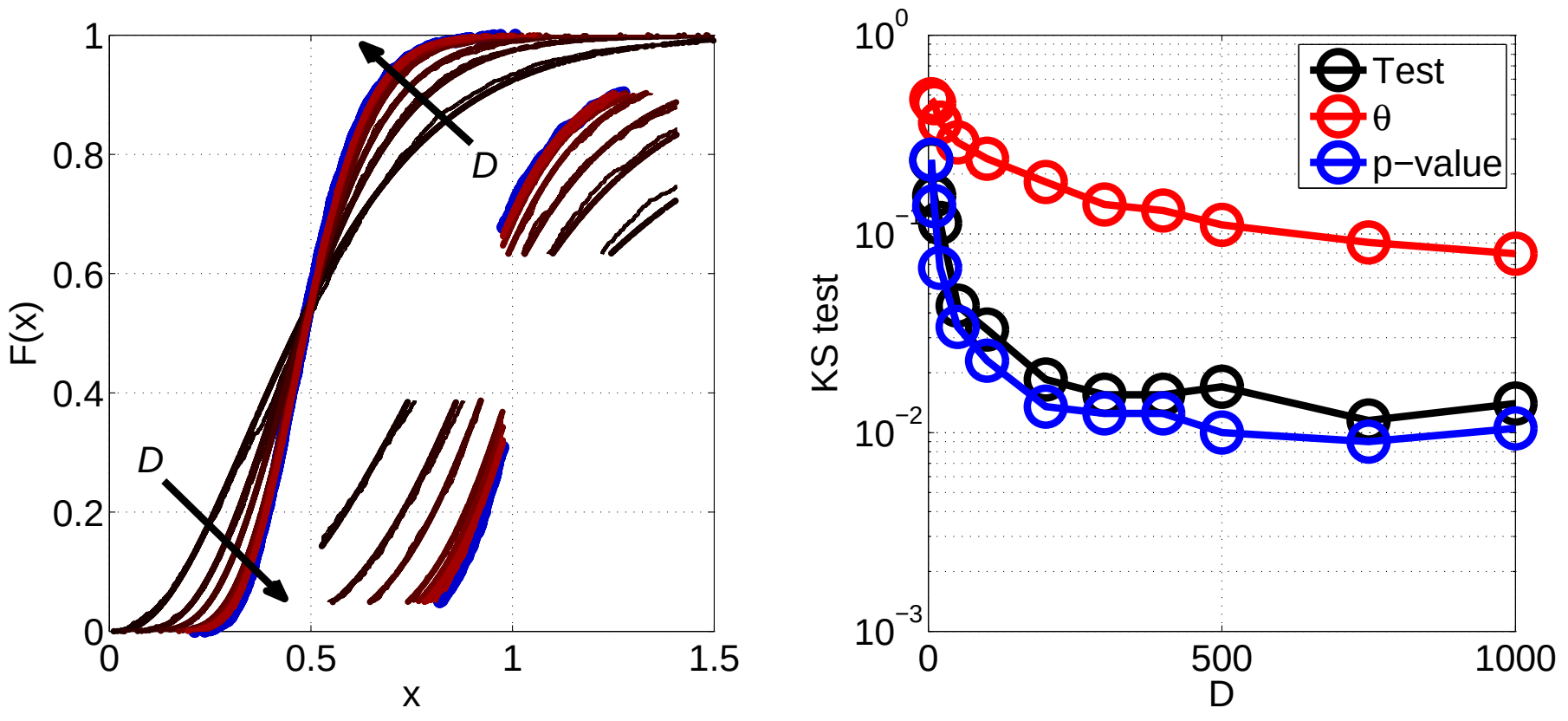

Figure 1: Comparison of HSIC and RHSIC. Left: Cumulative distribution function (cdf) of HSIC under $\mathrm{H}_0$ for $n = 100$ obtained empirically using 2000 independent draws of HSIC (blue line) and approximated using the two-parameter Gamma distribution for different numbers $D$ of random features for RHSIC (red-to-brownish lines), see zooomed insets for details on the approximations close to the low/high cdf values. Right: We show the Kolmogorov-Smirnov (KS) test between the distributions of the test statistic, the threshold parameter, and the $p$-value of HSIC and RHSIC as a function of $D$. The thresholds $\theta$ correspond to the inverse cdf of the significance level $\alpha = 0.05$, and the $p$-value which corresponds to the estimated test.

Then, the threshold $\theta$ is computed through the inverse cumulative density function (cdf) of the $1 - \alpha$ value, where $\alpha$ is the adopted significance level (typically, $\alpha = 0.05$ or $\alpha = 0.01$). Two random variables are then considered dependent if $\widehat{\mathrm{HSIC}} \geq \theta$, and independent otherwise. Alternatively, one can directly compute the HSIC $p$-value from the HSIC estimate and its cdf to test dependence. The $p$-value represents the probability of obtaining a result at least as extreme as the actually observed, assuming that the null hypothesis is true. Figure 1 shows the convergence (in Kolmogorov-Smirnov test terms) of the Gamma distribution of RHSIC to that of HSIC for different numbers of random features $D$. We also show convergence of thresholds $\theta$ and the corresponding

[1] The null hypothesis $\mathrm{H}_0$ is accepted if the test is lower than the threshold. The Type I error is defined as the probability of rejecting $\mathrm{H}_0$ based on the observed sample, despite $\mathbf{x}$ and $\mathbf{y}$ being independent. Conversely, the Type II error is the probability of accepting the null hypothesis when the underlying variables are dependent. The level $\alpha$ of a test is an upper bound on the Type I error, and is used to set the test threshold.

$p$-values. All measures suggest a fast convergence for few random features (note the log-scale).

# 3 The HSIC Sensitivity Maps

HSIC has demonstrated excellent capabilities to detect dependence between random variables but, as for any kernel method, the learned relations are hidden behind the kernel feature mapping. Visualization and geometrical interpretation of kernel methods in general and kernel dependence estimates in particular is an important issue in machine learning. To address this issue, we derive Sensitivity Maps (SMs) of both HSIC and RHSIC kernel dependence measures.

## 3.1 Sensitivity analysis and maps

A general definition of the sensitivity map was originally introduced in (Kjems et al., 2002). Let us define a function $f(\mathbb{z}) : \mathbb{R}^d \to \mathbb{R}$ parametrized by $\mathbb{z} = [z_1, \ldots, z_d]$. The sensitivity map for the $j$th feature, $z_j$, is the expected value of the squared derivative of the function (or the log of the function) with respect the arguments. Formally, let us define the sensitivity of $j$th feature as

$$s_j = \int_{\mathcal{Z}} \left( \frac{\partial f(\mathbb{z})}{\partial z_j} \right)^2 p(\mathbb{z}) \mathrm{d}\mathbb{z}, \tag{6}$$

where $p(\mathbb{z})$ is the probability density function over the input $\mathbb{z} \in \mathcal{Z}$. Intuitively, the objective of the sensitivity map is to measure the changes of the function $f(\mathbb{z})$ in the $j$th direction of the input space. In order to avoid the possibility of cancellation of the terms due to its signs, the derivatives are typically squared, even though other unambiguous transformations like the absolute value could be equally applied. Integration gives an overall measure of sensitivity over the observation space $\mathcal{Z}$. The *empirical sensitivity map* approximation to Eq. (6) is obtained by replacing the integral with a summation over the available $n$ samples

$$s_j \approx \frac{1}{n} \sum_{i=1}^{n} \left. \frac{\partial f(\mathbb{z})}{\partial z_j} \right|_{\mathbb{z}_i}^2, \tag{7}$$

which can be collectively grouped to define the *sensitivity vector* as $\mathbf{s} := [s_1, \ldots, s_d]$.

## 3.2 Sensitivity maps for the HSIC

The previous definition of SMs is limited in some aspects: 1) it cannot be directly applied to general functions $f$ depending on matrices or tensors, and 2) it allows estimating feature relevances $s_j$ only, discarding the individual samples relevance. The former is a severe limitation in particular for HSIC as it directly deals with data matrices. The latter limitation affects all type of functions including HSIC, which could be useful in applications on outlier detection or sample selection, just to name a few. We will derive the HSIC with respect to both the input samples and features simultaneously to address both shortcomings.

Let us now define $f :=$ HSIC, and replace the gradient $\frac{\partial f(\mathbb{z})}{\partial z_j}$ with $\frac{\partial f(\mathbb{Z})}{\partial Z_{ij}}$. Note that point-wise HSIC is parametrized by $\mathbb{z} \sim \mathbf{x}, \mathbf{y}|\sigma$, while a matrix parametrization reduces to $\mathbb{Z} \sim \mathbf{X}, \mathbf{Y}|\sigma$. In order to compute the HSIC sensitivity maps, we derive HSIC w.r.t. input data matrix entries $X_{ij}$ and $Y_{ij}$. By applying the chain rule, and first-order derivatives of matrices, we obtain:

$$S_{ij}^x := \frac{\partial \text{HSIC}}{\partial X_{ij}} = -\frac{2}{\sigma^2 n^2} \text{Tr}\left(\mathbf{H}\mathbf{K}_y\mathbf{H}(\mathbf{K}_x \circ \mathbf{M}_j\right), \tag{8}$$

where for a given $j$th feature, entries of the corresponding matrix $\mathbf{M}_j$ are $M_{ik} = X_{ij} - X_{kj}$ $(1 \leq i, k \leq n)$, and the symbol $\circ$ is the Hadamard product between matrices. After operating similarly for $Y_{ij}$, we obtain the corresponding expression:

$$S_{ij}^y := \frac{\partial \text{HSIC}}{\partial Y_{ij}} = -\frac{2}{\sigma^2 n^2} \text{Tr}\left(\mathbf{H}\mathbf{K}_x\mathbf{H}(\mathbf{K}_y \circ \mathbf{N}_j)\right), \tag{9}$$

where entries of matrix $\mathbf{N}_j$ are $N_{ik} = Y_{ij} - Y_{kj}$, and we assumed the SE kernel for both variables[2].

It is worth noting that, even though one could be tempted to use each sensitivity map independently, the solution is a *vector field*, and we should treat the sensitivity map jointly. To do this we define the *total sensitivity map* for all features and samples as $\mathbf{S} := [\mathbf{S}^x, \mathbf{S}^y] \in \mathbb{R}^{n\times(d_x+d_y)}$. From $\mathbf{S}$ we can compute the *empirical sensitivity map* per either feature or sample by integration in the corresponding domain, whose empirical estimates are respectively $s_i \approx \frac{1}{d}\sum_{j=1}^{d} S_{ij}^2$ and $s_j \approx \frac{1}{n}\sum_{i=1}^{n} S_{ij}^2$. This complementary view of the sensitivity reports information about the directions most impacting the dependence estimate, and allows a quantitive geometric evaluation of the measure.

Note that both HSIC and its SM give raise to closed-form solutions, just involving simple matrix multiplication and a trace operation. However, HSIC and its sensitivity maps scale quadratically with the number of examples $n$ since the involved matrices, both the kernel matrices and the centering matrix are $n \times n$. This makes both HSIC and its sensitivity map infeasible with moderate to large datasets. Next, we propose the corresponding SM for RHSIC to improve its computational efficiency, while still retaining the desireable simplicity and closed-form properties.

## 3.3 Sensitivity maps for the RHSIC

The SM for RHSIC can be derived as follows. Components of the sensitivity map readily follow from applying the chain rule, and first-order derivatives of matrices,

$$\begin{aligned}\hat{S}_{ij}^x = \frac{\partial \text{RHSIC}}{\partial X_{ij}} &= \frac{2}{n^2}\Re(\text{Tr}(\tilde{\mathbf{Z}}_y\tilde{\mathbf{Z}}_y^\top\tilde{\mathbf{Z}}_x(\tilde{\mathbf{Z}}_x^\top \circ \mathbb{i}\mathbf{W}_x\mathbf{J}_{ji}))) \\ &= \frac{2}{n^2}\Re(\text{Tr}(\tilde{\mathbf{Z}}_y^\top\tilde{\mathbf{Z}}_x(\tilde{\mathbf{Z}}_x^\top \circ \mathbb{i}\mathbf{W}_x\mathbf{J}_{ji})\tilde{\mathbf{Z}}_y)),\end{aligned} \tag{10}$$

[2]Other positive definite universal kernels could be equally adopted, but that requires deriving the sensitivity maps for the specific kernel. In addition, as seen in Section 3.3, random features approximation works for shift invariant kernels only.

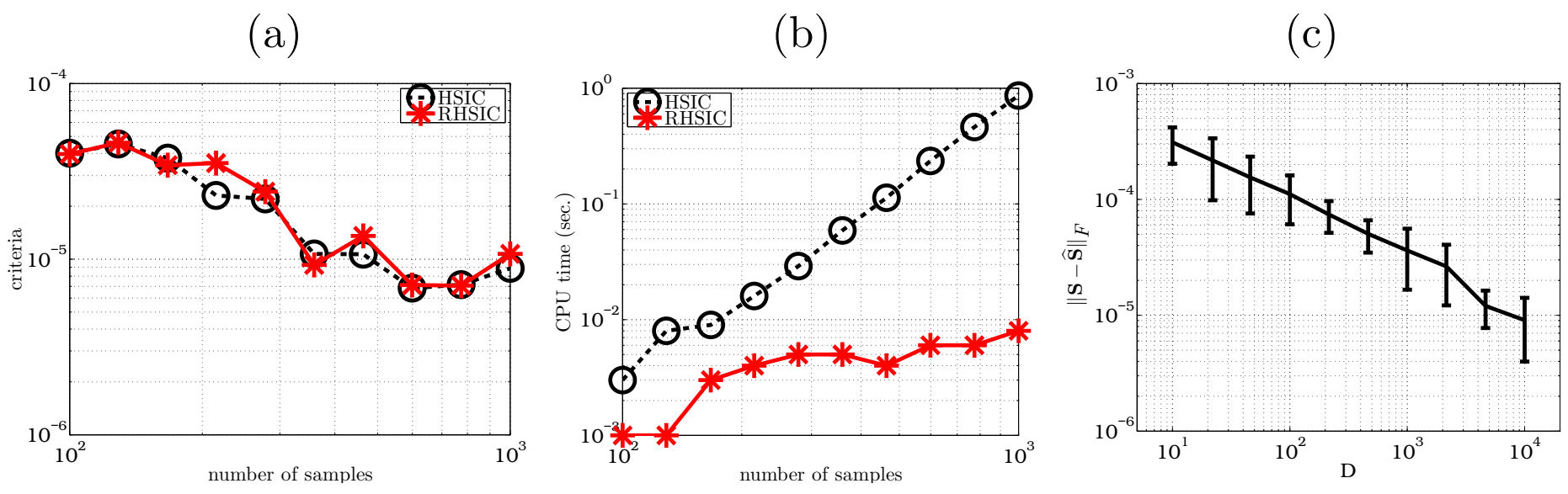

Figure 2: Empirical results of HSIC and RHSIC (a) criteria values, (b) computational running time and (c) sensitivity convergence.

where matrix $\mathbf{J}_{ji}$ is the single-entry matrix with entry 1 at $(i, j)$ and zero elsewhere. After operating similarly for $\mathbf{y}$, we obtain the expression:

$$\begin{aligned}\hat{S}^y_{ij} = \frac{\partial \text{RHSIC}}{\partial Y_{ij}} =& \frac{2}{n^2}\Re(\text{Tr}(\tilde{\mathbf{Z}}_x\tilde{\mathbf{Z}}_x^\top\tilde{\mathbf{Z}}_y(\tilde{\mathbf{Z}}_y^\top \circ \mathbb{i}\mathbf{W}_y\mathbf{J}_{ji})))\\ =& \frac{2}{n^2}\Re(\text{Tr}(\tilde{\mathbf{Z}}_x^\top\tilde{\mathbf{Z}}_y(\tilde{\mathbf{Z}}_y^\top \circ \mathbb{i}\mathbf{W}_y\mathbf{J}_{ji})\tilde{\mathbf{Z}}_x)). \end{aligned} \tag{11}$$

In both cases we used the permutation property of the trace which yields a convenient way to compute the randomized sensitivities as the trace is computed over $D_y \times D_x$ matrices and big centering matrices are not involved. The Theorem A demonstrate the convergence of RHSIC to HSIC.

# 4 Experimental results

This section gives empirical evidence of the performance of the proposed RHSIC and the sensitivity maps in several problems of data visualization, dependence estimation, feature selection and causal inference from empirical data.

## 4.1 Empirical convergence of approximations

The first experiment studies RHSIC and its sensitivity map in terms of convergence and computational cost as a function of the number of samples. In all cases the kernel function is the SE and the kernel parameter was fixed always to average Euclidean distance between training data. Figure 2(a) compares HSIC and RHSIC empirical estimates as a function of the sample size. Figure 2(b) shows a similar comparison in terms of the computational time. Results shown are the average of 10 realizations. The samples are drawn i.i.d. from two unidimensional uniform variables in $[0, 1]$. The number of features was fixed to $D = 30$. The differences between original HSIC and its approximation (Fig. 2(a)) are relatively low. This gives high efficiency gains (Fig. 2(b)): with such low value of $D$, RHSIC can deal with far more samples than HSIC without increasing its complexity. Figure 2(c) shows the Frobenius norm of the sensitivity map error between HSIC and RHSIC for different values of $D$. Excellent convergence rates are obtained as a function of $D$.

## 4.2 Visualization of sensitivity maps

Figure 3 shows several standard measures of dependence (Pearson's $\rho$, Spearman's rank, Kendall's $\tau$, mutual information) compared to canonical HSIC and its randomized version RHSIC. We show dependence estimates for 14 different problems of variable association. RHSIC performs virtually identical to HSIC in all cases, at a fraction of time. In addition, we show the sensitivity maps for all problems as colored points (lighter reddish means higher sample-dependent sensitivity norm, $\|\mathbf{s}_i\|$). It can be noted that low sensitivity values are commonly related to conditional means in Gaussian associations (top row), extremes, joints and change/elbow points in nonlinearly related variables, and clustered/randomly located points in near independence cases.

## 4.3 Robustness of sensitivity maps

Sensitivity maps are useful to visualize the impact of samples and features in kernel methods in general and kernel dependence measures in particular. Here we study the behavior of the SMs under different noise sources and types of association. Figure 4 shows the effect of sensitivity maps in both linear and nonlinear associations, buried under either Gaussian homoscedastic or heteroscedastic noise sources. Observed data points $(x, y)$ (yellow dots) are displaced to maximize dependence $(x+s_1, y+s_2)$ (orange dots), and the overall sensitivity (length of the velocity vector) increases in higher and heteroscedastic noise regimes, which could be eventually useful for noise estimation and outlier detection.

## 4.4 Sensitivity maps for feature selection

This section illustrates the use of the sensitivity of HSIC for feature selection in a real challenging problem. In particular, we deal with a regression setting, trying to link satellite observations with *in-situ* measurements of chlorophyll content (Chl-a) of plants in the land cover. The data used in this study were obtained in two terrestrial campaigns in Barrax, Spain. The test area has a rectangular form and an extent of 5 km × 10 km, and is characterized by a flat morphology and large, uniform land-use units. The region consists of approximately 65% dry land and 35% irrigated land. We used a calibrated CCM-200 Chlorophyll Content Meter to estimate Chl-a. Simultaneously, we used satellite images from the CHRIS sensor. CHRIS measures over the visible/ near-infrared spectra from 400 to 1050 nm. For this study, we used CHRIS data in Mode 1 (62 bands, full spectral information) for the four campaign days, where *in situ* measurements of surface properties were measured in conjunction with the satellite overpass. The images were geometrically and atmospherically corrected. A total set of 135 measurements in the 62-dimensional feature space constitute the database. Further details can be found in Verrelst et al. (2012).

We compare the results of feature ranking with standard dependence measures. A simple filter strategy was followed in the experiment: we selected a subset of features optimizing each criterion, $n_f = \{10, 15, 20\}$, and the quality of the selection was evaluated using the prediction goodness of fit measured as the Pearson's correlation coefficient, $\rho$, obtained by a Gaussian Process Regression (GPR) model. Figure 5(a) shows the obtained ranking results. The correlation coefficient is the result of averaging 100 realizations with different data partitions. The 95% confidence intervals are shown as well. It is noted that the HSIC-based criteria excel over correlation-based criteria and mutual information, regardless of the number of selected features. The gain obtained by the sensitivity measures is very noticeable, both numerically (around +7% in all cases) and statistically (as suggested by multi-way ANOVA on both bias and accuracy, $p < 0.01$). A further analysis of the selected features was revealing. Figure 5(b) shows the top 10 ranked features by the different criteria. It becomes clear that most of the methods focus on spectral bands (features) related to leaf pigments (600-700 nm), which have an obvious physical interpretation. However, HSIC and the proposed sensitivity criteria tend to select spectral channels associated with cell structure too, thus preferring spectral regions with subtle characteristics and highly aligned with chlorophyll content changes. Additionally, HSIC sensitivity criteria explore channels beyond those regions, which result in more expressive power and very useful to improve regression results.

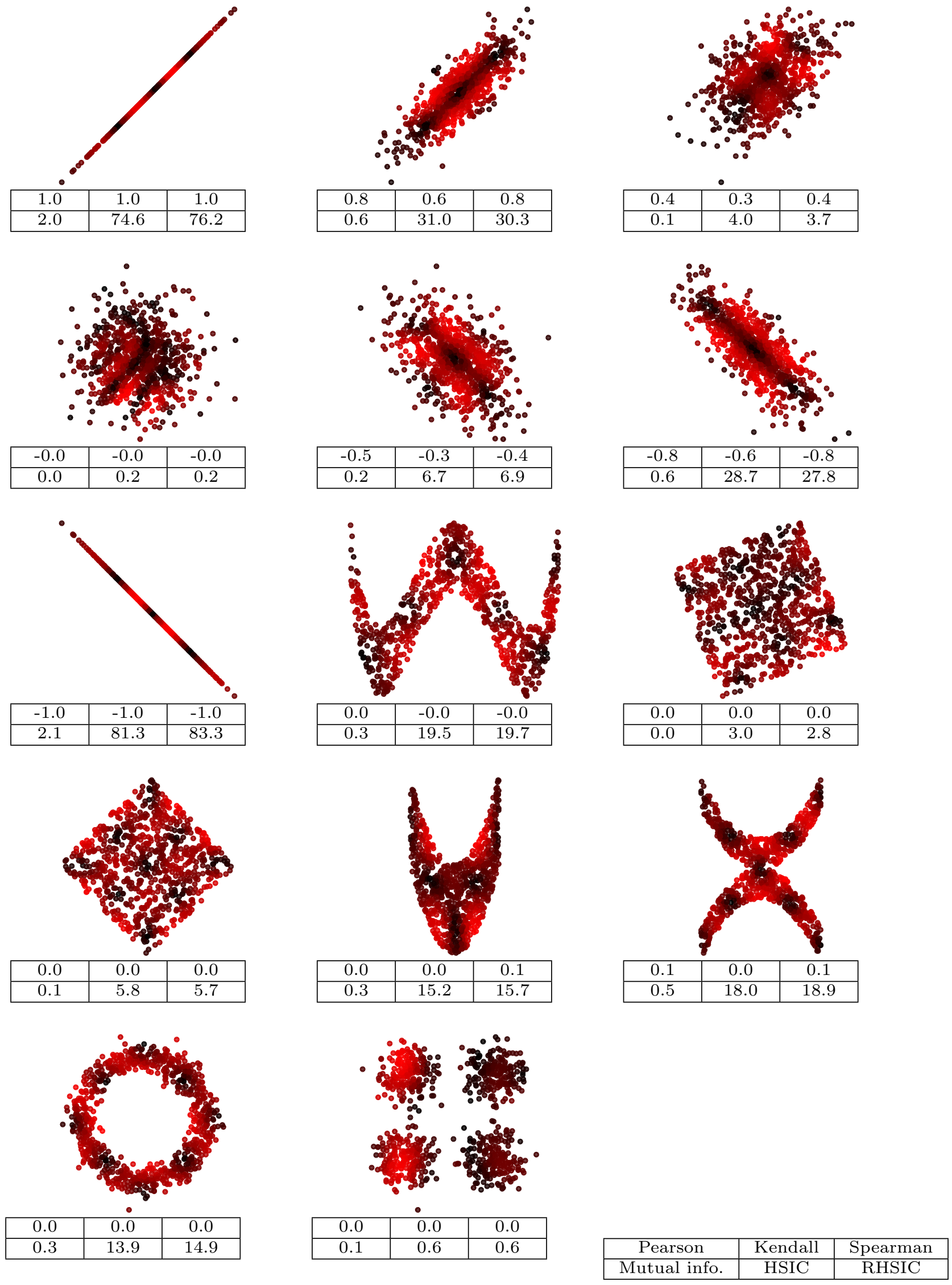

Figure 3: Results for dependence estimation in 14 toy examples.Each table gives Pearson, Kendall, Spearman (top, left-to-right) and mutual information, HSIC, RHSIC (bottom, left-to-right). We highlight the sensitivity norm per example (lighter reddish means higher sensitivity).

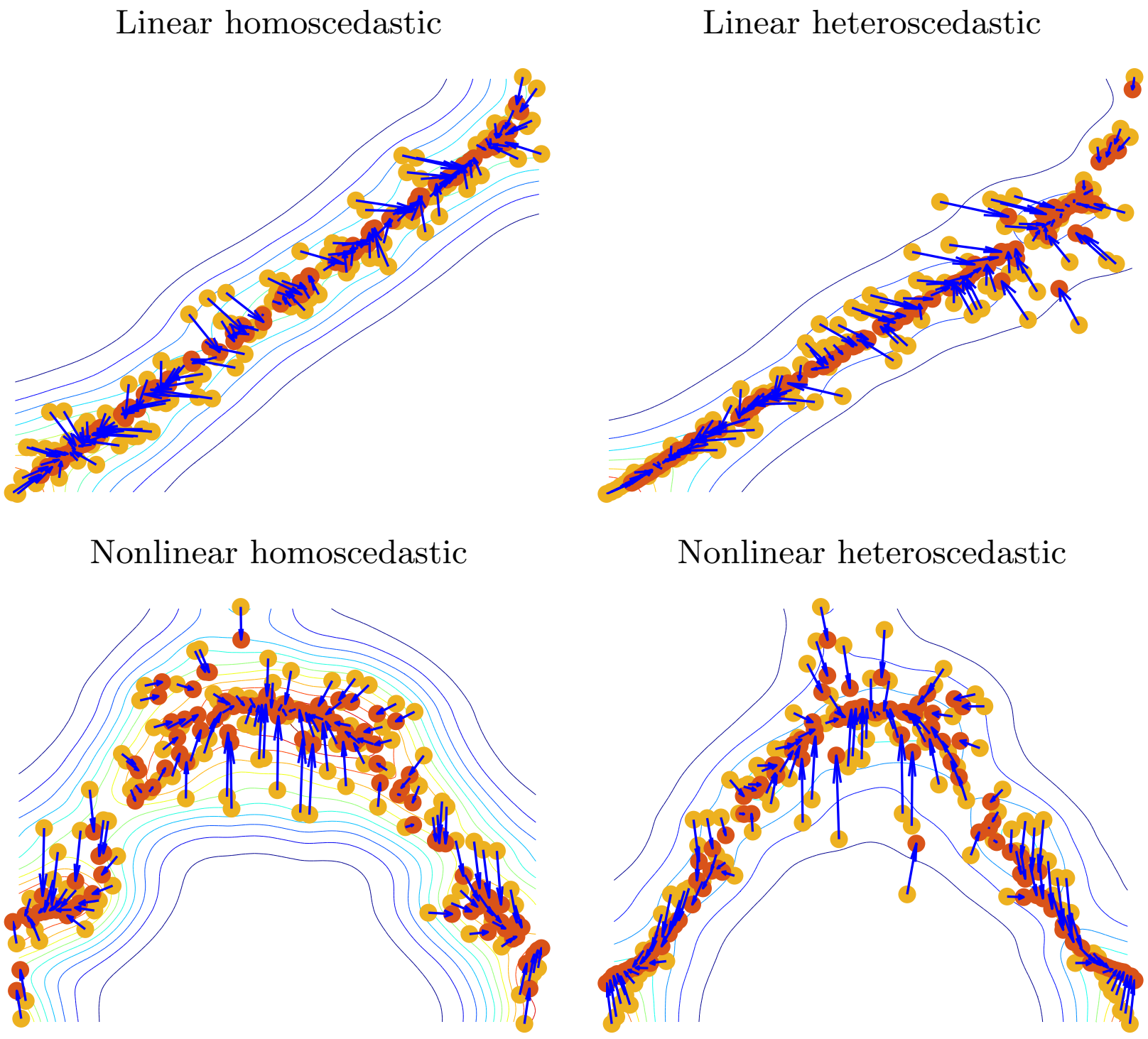

Figure 4: HSIC sensitivity maps in both linear and nonlinear with either homoscedastic or heteroscedastic noise sources.

### 4.5 Sensitivity maps for causal inference

Establishing causal relations between random variables from empirical data is perhaps the most important challenge in today's Science. In this experiment, we compare HSIC and RHSIC as dependence measures for causal inference. To this end, we follow the approach in Hoyer et al. (2008) to discover causal association between variables $x$ and $y$. The methodology performs nonlinear regression from $x \to y$ (and vice versa, $y \to x$) and assesses the independence of the forward, $r_f = y - f(x)$, and backward residuals, $r_b = x - g(y)$, with the input variable $y$ and $x$, respectively. The statistical significance of the independence test tells the right direction of causation. Essentially, the framework exploits nonlinear, non-parametric regression to assess the plausibility of the causal link between two random variables in both directions: statistically significant residuals in just one direction indicate the true data-generating mechanism.

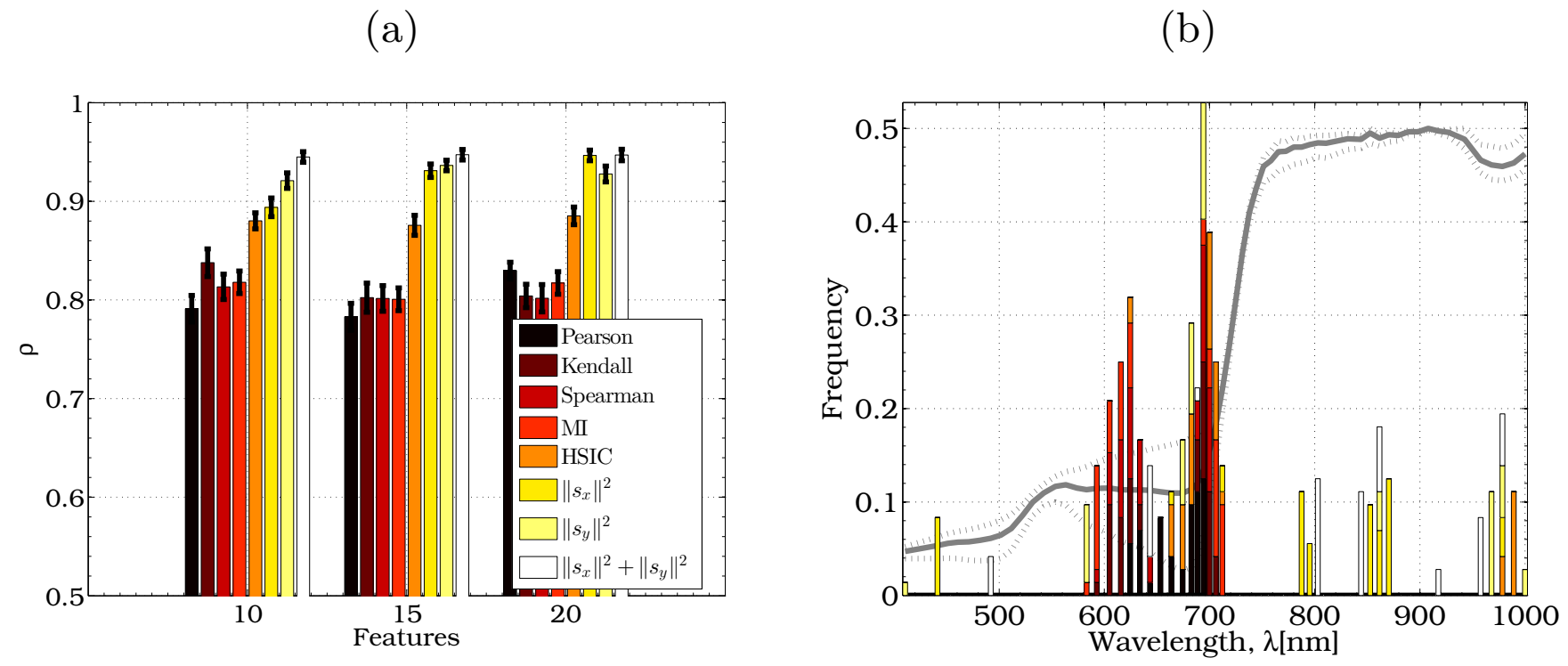

Figure 5: Feature selection results: (a) Pearson's correlation coefficient of the GPR models (averaged over 100 realizations) for different number of selected features and criteria; (b) Top 10 selected features by each dependence estimate (the averaged spectra and the $\pm$ standard deviation is plotted for reference of the spectral shape of species in the area).

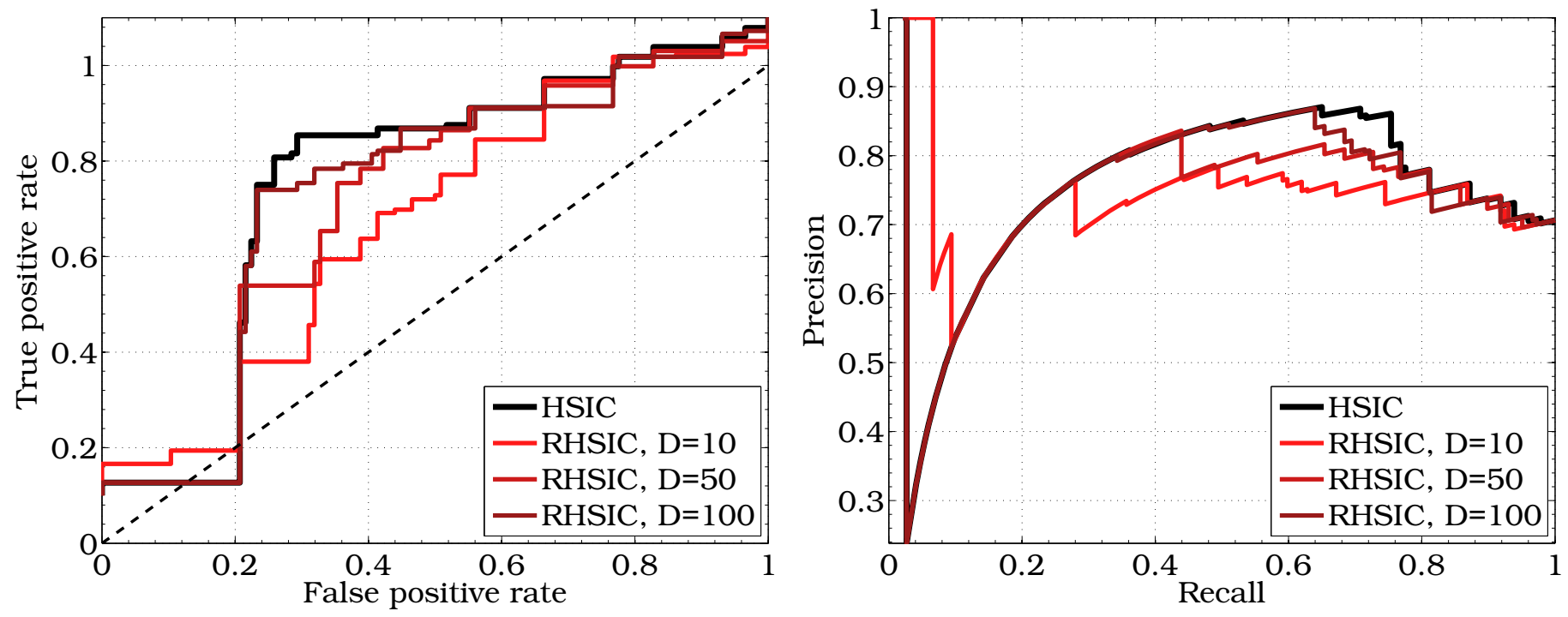

Figure 6: ROC (left) and Precision-Recall (right) curves for the causality problem using $n = 2000$ for training and different number of random features $D$.

We used Version 1.0 of the CauseEffectPairs (CEP) collection. The database contains 100 pairs of random variables along with the right direction of causation (ground truth). Data has been collected from 37 different data sets from various domains of application, from biology, climate science and economics, just to name a few. The CEP data are publicly available at `https://webdav.tuebingen.mpg.de/cause-effect/`. More information about the dataset and an excellent up-to-date review of causal inference methods is available in Mooij et al. (2016). We conducted experiments in 95 out of the 100 pairs that have one-dimensional variables.

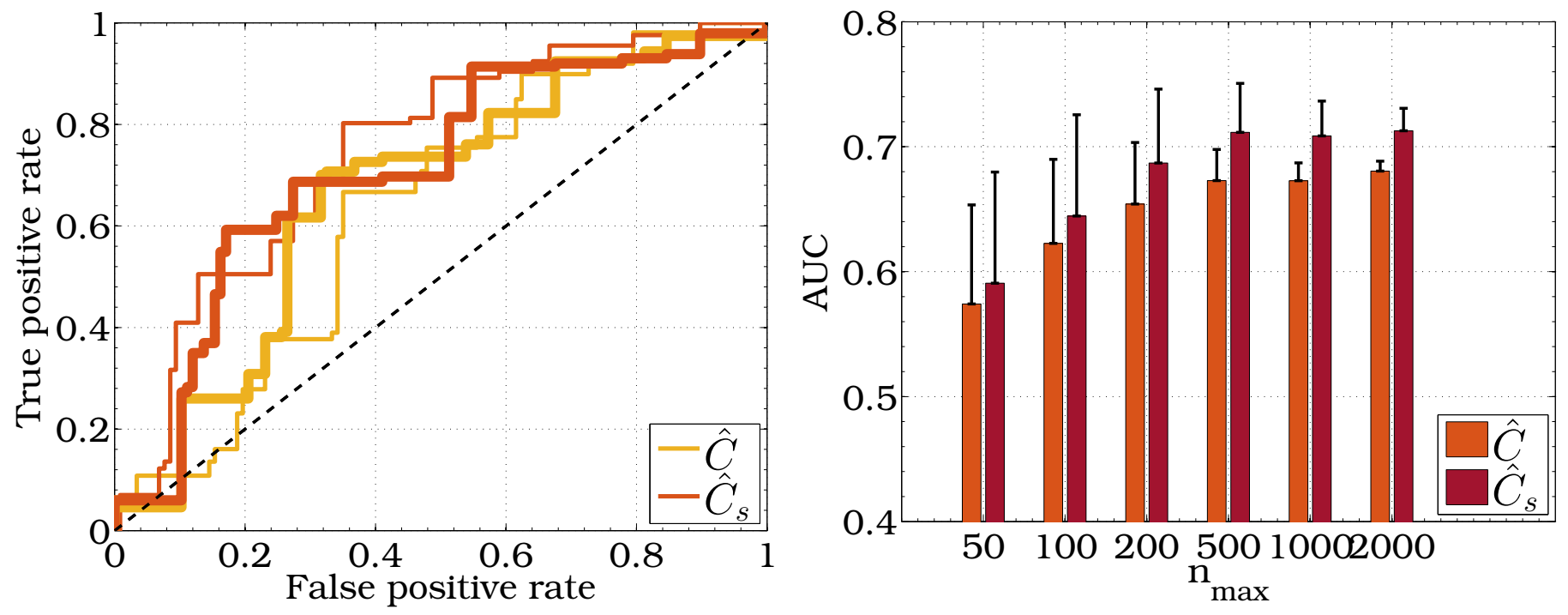

Figure 7: ROC and corresponding AUCs in the CEP causality problems dataset for $n_{max} = 200$ (thin lines) and $n_{max} = 2000$ (thick lines) in every problem.

The experimental setting is as follows: we used random forests (RFs)[3] for both forward and backward models, and then used HSIC and RHSIC to estimate independence of the corresponding residuals to rank the decisions $\hat{D}$ following Hoyer et al. (2008). The final causal direction score was simply defined as the difference in test statistic between both models

$$\hat{C} := \text{HSIC}(x, r_f) - \text{HSIC}(y, r_b).$$

This is a particular form of 'ranked-decision' setting that needs to account for the bias introduced by pairs coming from the same problem: it is customary to down-weight the precision for every decision threshold in the curves (e.g. four related problems receive 0.25 weights in the decision)[4]. We first aimed at assessing the performance of RHSIC as a function of $D$, and fixed the maximum number of training examples to $n_{max} = 2000$ in all problems. We computed Receiver-Operating Curve (ROC) and their corresponding areas under the curves (AUCs). Results showed that (1) RHSIC converges to HSIC as $D$ increases and for all decision rates, as expected; and (2) low values of $D$ resulted in higher (lower) true positive rates for low (high) levels of false positive rates. The AUC-ROC results for HSIC achieved AUC=0.6562, while RHSIC yielded AUC={0.5704, 0.6111, 0.6344} for $D = \{10, 50, 100\}$ respectively. We can conclude that convergence of RHSIC to the full HSIC estimates and the corresponding causal detection curves is fast, and does not need too many features to achieve similar results.

[3] We also tried Gaussian process regression as in Mooij et al. (2016) but results were less accurate and more computationally demanding.

[4] The MATLAB function `perfcurve` can produce such (weighted) ROC and PRC curves and the estimated weighted AUC.

Next we assessed performance when working with few examples $n$. This situation dramatically impacts regression models performance, both in terms of the regression accuracy and the dependence estimation. The problem can be alleviated by including the sensitivity maps in the same scheme to account for the 'metric' of the dependence measure. In particular, we here propose the use of an alternative criterion for association:

$$\hat{C}_s := (S_b^y + S_b^r) - (S_f^x + S_f^r),$$

where subscripts $f$ and $b$ stand for the forward and backward directions, and the superscripts refer to the sensitivities of either the observations $x$ and $y$, or the corresponding residuals. Similar criteria has been recently presented in the literature, yet focusing on the derivative of the underlying function, instead of the derivative of the dependence estimate itself (Daniusis et al., 2012).

We evaluate $\hat{C}$ and $\hat{C}_s$ criteria by limiting the maximum number of training samples in each problem, $n_{max} = \{50, 100, 200, 500, 2000\}$. Results were averaged over 10 realizations. Figure 7 shows the ROCs for both criteria, and the AUC under the curves as a function of $n_{max}$. The proposed sensitivity-based criterion consistently performed better than the standard approach using HSIC alone.

# 5 Conclusions

In this paper we introduced the sensitivity analysis for kernel dependence measures and particularized the study for the familiar Hilbert-Schmidt independence criterion. The sensitivity maps allow to study the measure in geometric terms, and analyze the relative relevance of features and samples most impacting dependence between variables. The analysis has some good properties, namely its empirical estimate is simple to compute in closed-form and was shown to provide side information useful for data visualization, feature ranking and causal inference.

Noting that HSIC has an important computational cost, we introduced a randomized version based on the Bochner's theorem to approximate the involved kernels in the canonical HSIC with projections on random features. The power of the RHSIC measure scales favorably with the number of samples, and it approximates HSIC and the sensitivity maps efficiently. We provided convergence bounds for both the measure and the sensitivity maps. We successfully illustrated the performance of HSIC, RHSIC and their sensitivity maps in synthetic and real machine learning data problems. We would like to note that the use of random features for HSIC approximation is not incidental. The proposed RHSIC improves the computational efficiency of standard HSIC and still allows us to estimate the sensitivity in closed form.

It does not escape our notice that the sensitivity maps can find further application to identify samples and features most impacting the measure, thus being potentially useful for anomaly detection, finding sparse representations, and noise estimation. These issues will be subject of further research.

# A Convergence bounds of randomized HSIC

Here we give guarantees of the convergence of RHSIC to HSIC. The solution of HSIC roughly involves multiplication of (centered) kernel matrices $\mathbf{K}_x$ and $\mathbf{K}_y$, which are approximated by $\hat{\mathbf{K}}_x = \hat{\mathbf{Z}}_x\hat{\mathbf{Z}}_x^\top$ and $\hat{\mathbf{K}}_y = \hat{\mathbf{Z}}_y\hat{\mathbf{Z}}_y^\top$, using $D_x$ and $D_y$ random features, respectively, and $\hat{\mathbf{Z}} = \mathbf{ZH}$ represent centered matrices. Our aim is to give a bound on the approximation error to a product of kernel matrices from products of random projection matrices. First we recall the Hermitian Matrix Bernstein theorem, which is then used to derive the bound for RHSIC.

**Theorem** (Matrix Bernstein, Mackey et al. (2014))**.** *Let $\mathbf{Z}_1, \ldots \mathbf{Z}_m$ be independent $n \times n$ random matrices. Assume that $\mathbb{E}[\mathbf{Z}_i] = 0$ and that $\|\mathbf{Z}_i\| \leq R$. Define the variance parameter $\sigma^2 := \max\{\|\sum_i \mathbb{E}[\mathbf{Z}_i^\top \mathbf{Z}_i]\|, \|\sum_i \mathbb{E}[\mathbf{Z}_i\mathbf{Z}_i^\top]\|\}$. Then, for all $t \geq 0$,*

$$\mathbb{P}\left(\left\|\sum\nolimits_i \mathbf{Z}_i\right\| \geq t\right) \leq 2n \exp\left(\frac{-t^2}{3\sigma^2 + 2Rt}\right)$$

*and*

$$\mathbb{E}\left\|\sum\nolimits_i \mathbf{Z}_i\right\| \leq \sqrt{3\sigma^2 \log(n)} + R\log(n).$$

**Theorem.** *Assume access to the datasets $\mathbf{X} \in \mathbb{R}^{n\times p}$, $\mathbf{Y} \in \mathbb{R}^{n\times q}$ and shift-invariant kernels $K_x$, $K_y$. Define the (centered) kernel matrices $(\mathbf{K}_x)_{ij} := K_x(\mathbf{x}_i, \mathbf{x}_j)$, $(\mathbf{K}_y)_{ij} := K_y(\mathbf{y}_i, \mathbf{y}_j)$ and their approximations $\hat{\mathbf{K}}_x = \tilde{\mathbf{Z}}_x\tilde{\mathbf{Z}}_x^\top$ and $\hat{\mathbf{K}}_y = \tilde{\mathbf{Z}}_y\tilde{\mathbf{Z}}_y^\top$ using $D_x$, $D_y$ random features, where $\tilde{\mathbf{Z}} = \mathbf{ZH}$ represent centered matrices, and $D := \min(D_x, D_y)$. Then, the $\ell_2$ approximation error can be bounded as*

$$\mathbb{E}\|\hat{\mathbf{K}}_x\hat{\mathbf{K}}_y - \mathbf{K}_x\mathbf{K}_y\| \leq \sqrt{\frac{3n^4\log(n)}{D}} + \frac{2n^2\log(n)}{D}, \tag{12}$$

*which shows a logarithmic trend with $D$, linear in the log-scale $\mathcal{O}(D^{-1/2})$, and $n\log(n)$ trend with the number of training examples.*

*Proof.* We follow a similar derivation to Lopez-Paz et al. (2014) for randomized nonlinear CCA. The total error matrix can be decomposed as a sum of individual error terms, $\mathbf{E} = \sum_{i=1}^{D_y} \mathbf{E}_i$, which are defined as $\mathbf{E}_i = \frac{1}{D_y}\big(\hat{\mathbf{K}}_x\hat{\mathbf{K}}_y^{(i)} - \mathbf{K}_x\mathbf{K}_y\big)$. Since the $D_x + D_y$ random features are sampled i.i.d., and the data matrices for each domain are constant, the random matrices $\{\hat{\mathbf{K}}_x^{(1)}, \ldots, \hat{\mathbf{K}}_x^{(D_x)}, \hat{\mathbf{K}}_y^{(1)}, \ldots, \hat{\mathbf{K}}_y^{(D_y)}\}$ are i.i.d. random variables. Hence, their expectations factorize, $\mathbb{E}\left[\mathbf{E}_i\right] = \frac{1}{D_y}\big(\mathbb{E}[\hat{\mathbf{K}}_x]\mathbf{K}_y - \mathbf{K}_x\mathbf{K}_y\big)$, where we used $\mathbb{E}[\hat{\mathbf{K}}_y^{(i)}] = \mathbf{K}_y$. The deviation of the individual error matrices from their expectations is $\mathbf{Z}_i = \mathbf{E}_i - \mathbb{E}\left[\mathbf{E}_i\right] = \frac{1}{D_y}\big(\hat{\mathbf{K}}_x\hat{\mathbf{K}}_y^{(i)} - \mathbb{E}[\hat{\mathbf{K}}_x]\mathbf{K}_y\big)$. Now we can apply Hölder's condition twice after using the triangle inequality on the norm, and Jensen's inequality on the expected values and obtain a bound of the error matrices, $R$:

$$\|\mathbf{Z}_i\| = \frac{1}{D_y}\|\hat{\mathbf{K}}_x\hat{\mathbf{K}}_y^{(i)} - \mathbb{E}[\hat{\mathbf{K}}_x]\mathbf{K}_y\| \leq \frac{B(B + \|\mathbf{K}_y\|)}{D_y} \leq \frac{2n^2}{D_y},$$

where $B$ is a bound on the norm of the randomized feature map, $\|\mathbf{z}\|^2 \leq B$. The variance is defined as $\sigma^2 := \max\{\|\sum_{i=1}^{D_y}\mathbb{E}[\mathbf{Z}_i\mathbf{Z}_i^\top]\|, \|\sum_{i=1}^{D_y}\mathbb{E}[\mathbf{Z}_i^\top\mathbf{Z}_i]\|\}$. Let

us expand the individual terms in the (first) summand:

$$\begin{aligned}\tilde{\mathbf{Z}}_i^\top \tilde{\mathbf{Z}}_i &= \frac{1}{D_y^2}\Big(\hat{\mathbf{K}}_y^{(i)}\hat{\mathbf{K}}_x^2\hat{\mathbf{K}}_y^{(i)} + \mathbf{K}_y\mathbb{E}[\hat{\mathbf{K}}_x]^2\mathbf{K}_y \\ &\quad -\hat{\mathbf{K}}_y^{(i)}\hat{\mathbf{K}}_x\mathbb{E}[\hat{\mathbf{K}}_x]\mathbf{K}_y - \mathbb{E}[\hat{\mathbf{K}}_x]\mathbf{K}_y\hat{\mathbf{K}}_x\hat{\mathbf{K}}_y^{(i)}\Big),\end{aligned} \tag{13}$$

and now taking the norm of the expectation, and using Jensen's inequality, we obtain $\left\|\mathbb{E}[\tilde{\mathbf{Z}}_i^\top\tilde{\mathbf{Z}}_i]\right\| \leq \frac{B^2\|\mathbf{K}_y\|^2}{D^2}$, which is the same for $\|\mathbb{E}[\tilde{\mathbf{Z}}_i\tilde{\mathbf{Z}}_i^\top]\|$, and therefore the worst-case estimate of the variance is $\sigma^2 \leq \frac{B^2\|\mathbf{K}_y\|^2}{D_y}$. The bound can be readily obtained using the Hermitian matrix Bernstein inequality, and the fact that random features and kernel evaluations are upper-bounded by 1, and thus both $B$ and $\|\mathbf{K}\|$ are upper-bounded by $n$. □